%% file: main.tex
\documentclass[runningheads]{llncs}
\usepackage{graphicx}
\usepackage{multicol}
\usepackage{multirow}
\usepackage{amsmath}
\usepackage{amssymb}
\usepackage{booktabs}
\usepackage{hyperref}
\usepackage{style}
\usepackage[normalem]{ulem}
\usepackage{xcolor}
\usepackage{centernot}
\usepackage{mathtools}
\usepackage{stmaryrd}
\usepackage{placeins}
\newcommand{\rpm}{\sbox0{$1$}\sbox2{$\scriptstyle\pm$}
 \raise\dimexpr(\ht0-\ht2)/2\relax\box2 }

\newcommand{\LTAE}{\text{L-TAE}}
\newcommand{\MLP}{\text{MLP}}

\begin{document}
\title{Lightweight Temporal Self-Attention  \\ for Classifying Satellite Images Time Series}
\titlerunning{Lightweight Temporal Attention}
%
\author{Vivien Sainte Fare Garnot \and
Loic Landrieu
}
\authorrunning{Sainte Fare Garnot V. and Landrieu L.}
%
\institute{ LASTIG, ENSG, IGN,
   Univ Gustave Eiffel,\\
  F-94160 Saint-Mande, France \\
\url{https://www.umr-lastig.fr/}}
\maketitle              
\begin{abstract}
The increasing accessibility and precision of Earth observation satellite data offers considerable opportunities for industrial and state actors alike. This calls however for efficient methods able to process time-series on a global scale.
Building on recent work employing multi-headed self-attention mechanisms to classify remote sensing time sequences, we propose a modification of the Temporal Attention Encoder of Garnot \etal \cite{garnot2019satellite}.
In our network, the channels of the temporal inputs are distributed among several compact attention heads operating in parallel. Each head extracts highly-specialized temporal features which are in turn concatenated into a single representation.
Our approach outperforms other state-of-the-art time series classification algorithms on an open-access satellite image dataset, while using significantly fewer parameters and with a reduced computational complexity.
\end{abstract}
%
%
%
\section{Introduction}
\input{intro}
\section{Method}
\input{method}
\section{Numerical Experiment}
\input{experiments}
\section{Conclusion}
We presented a new lightweight network for embedding sequences of observations such as satellite time-series. Thanks to a channel grouping strategy and the definition of the master query as a trainable parameter, our proposed approach is more compact and computationally efficient than other attention-based architectures.
Evaluated on an open-access satellite dataset, the L-TAE performs better than state-of-the-art approaches, with significantly fewer parameters and a reduced computational load, opening the way for continent-scale automated analysis of Earth observation.

Our implementation of the L-TAE can be accessed in the open-source repository: \url{github.com/VSainteuf/lightweight-temporal-attention-pytorch}.
%
%
%
\section*{Acknowledgments}
This research was supported by the AI4GEO project: \url{http://www.ai4geo.eu/} and the French Agriculture Paying Agency (ASP).

\section*{Appendix}

\input{appendix}
\FloatBarrier
\bibliographystyle{splncs04}
\bibliography{tae.bib}
\end{document}

%% file: intro.tex
Time series of remote sensing data, such as satellites images taken at regular intervals, provide a wealth of useful information for Earth monitoring. However, they are also typically very large, and their analysis is resource-intensive. For example, the Sentinel satellites gather over $25$ Tb of data every year in the EU. While exploiting the spatial structure of the data poses a challenge on its own, we focus in this paper on the efficient extraction of discriminative temporal features from sequences of spatial descriptors.

Among the many possible approaches to handling time-series of remote sensing data, one can concatenate observations in the temporal dimension \cite{kussul2016parcel}, use temporal statistics \cite{pelletier2016assessing}, histograms \cite{bailly2015dense}, time-kernels \cite{tavenard2017efficient}, or shapelets \cite{ye2009time}. Probabilistic graphical models such as Conditional Random Fields can also be used to exploit the temporal structure of the data \cite{bailly2018crop}.

Deep learning-based methods are particularly well-suited for dealing with the large amount of data collected by satellite sensors. Neural networks can either model the temporal dimension independently of the spatial dimensions with recurrent Neural Networks \cite{garnot2019time} or one-dimensional convolutions \cite{pelletier2019temporal}, or jointly with convolutional recurrent networks \cite{russwurm2018convolutional} or 3D convolutions \cite{ji20183d}.

More recently, the self-attention mechanism introduced by Vaswani \etal \cite{vaswani2017attention}, initially developed for Natural Language Processing (NLP), has been successfully used and adapted to remote sensing tasks \cite{russwurm2019self,garnot2019satellite}. In \secref{sec:selfattention}, we present these approaches and their differences in greater details.

 In this paper, we introduce the Lightweight Temporal Attention Encoder (L-TAE), a novel attention-based network focusing on memory and computational efficiency. Our approach is based on the Temporal Attention Encoder (TAE) of Garnot \etal \cite{garnot2019satellite}, with several modifications 
 meant to avoid redundant computations and parameters, while retaining a high degree of expressiveness and adaptability.
 We evaluate the performance of our approach on the open-access dataset Sentinel2-Agri \cite{garnot2019satellite}. With an equal parameter count, our algorithm outperforms all state-of-the-art competing methods in terms of precision and computational efficiency. Our method allows for efficient parameters usage, as our L-TAE outperforms TAEs with close to $10$ times the parameter count, as well as recurrent units over $300$ times larger.

%% file: method.tex
Throughout this section, we consider a generic input time series of length $T$ comprised of $E$-dimensional feature vectors $\mathbf{e} = [e^{(1)}, \cdots, e^{(T)}] \in \mathbb{R}^{E \times T}$. 
For example, such vectors can be a sequence of learned embeddings of super-spectral satellite images. 
\subsection{Multi-Headed Self-Attention}
\label{sec:selfattention}
In its original iteration \cite{vaswani2017attention}, self-attention---initially designed for text translation---consists of the following steps:
\begin{itemize}
    \item[](i) compute a triplet of key-query-value $k^{(t)}, q^{(t)}, v^{(t)}$ for each position $t$ of the input sequence with a shared linear layer applied to $e^{(t)}$,
    \item[](ii) compute attention masks representing the compatibility (dot-product) between the queries at each position and the keys corresponding to previous elements in the sequence,
    \item[](iii) associate to each position of the sequence an output defined as the sum of the previous values weighted by the corresponding attention mask. 
\end{itemize}
This process is done in parallel for $H$ different sets of independent parameters---or heads---whose outputs are then concatenated. This scheme allows each head to specialize in detecting certain characteristics of the feature vectors.

Ru{\ss}wurm \etal \cite{russwurm2019self} propose to apply this architecture to embed sequences of satellite observations by max-pooling the resulting sequence of outputs in the temporal dimension.
%
Garnot \etal \cite{garnot2019satellite} introduce the TAE, a modified self-attention scheme. First, they propose to directly use the input embeddings as values ($v^{(t)} = e^{(t)}$), taking advantage of the end-to-end training of the image embedding functions alongside the TAE. Additionally, they define a single master query $\hat{q}$ for each sequence, computed from the temporal average of the queries. This master query is compared to the sequence of keys to produce a single attention mask of dimension $T$ used to weight the temporal mean of values into a single feature vector. 
\subsection{Lightweight Attention}
\begin{figure}[t]
    \centering
    \includegraphics[width=\linewidth, trim=0cm 7cm .65cm 0cm , clip]{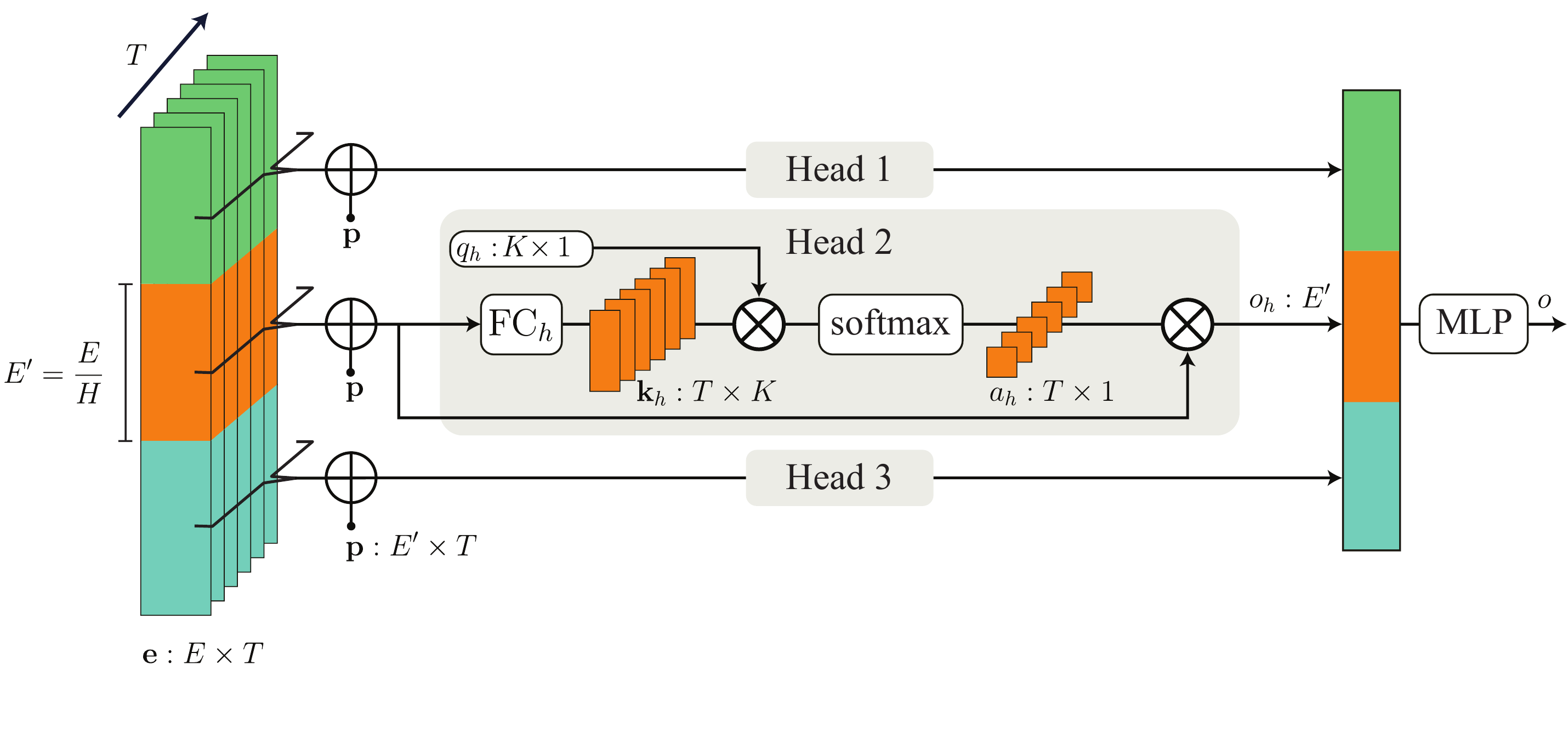}
    \caption{The proposed L-TAE module processing an input sequence $\mathbf{e}$ of $T$ vectors of size $E$, with $H=3$ heads and keys of size $K$.
    The channels of the input embeddings are distributed among heads. Each head uses a learnt query $\hat{q}_h$, while a linear layer FC$_h$ maps inputs to keys. The outputs of all heads are concatenated into a vector with the same size as the input embeddings, regardless of the number of heads. }
    \label{fig:pipeline}
\end{figure}
We build on this effort to adapt multi-headed self-attention to the task of sequence embedding. Our focus is on efficiency, both in terms of parameter count and computational load. 
%
\paragraph{Channel Grouping:} we propose to split the $E$ channels of the input elements into $H$ groups of size $E'=E/H$ with $H$ being the number of heads\footnote{$E$ and $H$ are typically powers of $2$ and $E>H$, ensuring that $E'$ remains integer.}, in the manner of Wu \etal \cite{wu2018group}. %
We denote by $e^{(t)}_h$ the groups of input channels for the $h$-th group of the $t$-th element of the input sequence \eqref{eq:hydra_split}.

We encode the number of days elapsed since the beginning of the sequence into an $E'$-dimensional positional vector $p$ of characteristic scale $\tau=1000$ \eqref{eq:hydra_pos}. In order for each head to access this information, $p$ is duplicated and added to each channel group.
Each head operates in parallel on its corresponding group of channels, thus accelerating the costly computation of keys and queries. This also allows for each head to specialize alongside its channel group, and avoid redundant operations between heads.
%
\paragraph{Query-as-Parameter:}
We define the $K$-dimensional master query $q_h$ of each head $h$ as a model parameter instead of the results of a linear layer. The immediate benefit is a further reduction of the number of parameters, while the lack of flexibility is compensated by the larger number of available heads.
\paragraph{Attention Masks:}
As a result, only the keys are obtained with a learned linear layer \eqref{eq:hydra_k}, while values are bypassed ($v^{(t)} = e^{(t)}$), and the queries are model parameters. The attention masks $a_h \in [0,1]^T$ of each head $h$ are defined as the scaled \emph{softmax} of the dot-product between the keys and the master query \eqref{eq:hydra_ah}.
The outputs $o_h$ of each heads are defined as the sum in the temporal dimension of the corresponding inputs weighted by the attention mask $a_h$ \eqref{eq:hydra_oh}. Finally, the heads outputs are concatenated into a vector of size $E$ and processed by a multi-layer perceptron $\MLP$ to the desired size \eqref{eq:hydra_out}.
%
In \figref{fig:pipeline}, we represent a schematic representation of our network. The different steps of the $\LTAE$  can also be condensed by the following operations, for $h = 1 \cdots H$ and $t = 1 \cdots T$:
\begin{align}
        e^{(t)}_h &=\left[e^{(t)}\left[(h-1)E' +i\right]\right]_{i=1}^{ E'} \label{eq:hydra_split}\\
        p^{(t)} &= \left[sin\left(\text{day}(t)/\tau^{\frac{i}{E'}}\right)\right]_{i=1}^{E'}\label{eq:hydra_pos}\\
        k^{(t)}_h &= \text{FC}_h (e^{(t)}_h + p^{(t)} )\label{eq:hydra_k}\\
        a_h &= \text{softmax} \left(  \frac{1}{\sqrt{K}}\left[q_{h} \cdot k^{(t)}_{h}\right]_{t=1}^T   \right) \label{eq:hydra_ah}\\
        o_{h} &=  \sum_{t = 1}^{T} a_h[t] \Pa{e^{(t)}_h + p^{(t)}} \label{eq:hydra_oh}\\
          o &= \MLP([o_{1}, \cdots, o_{H}])~. \label{eq:hydra_out}
\end{align}
\subsection{Spatio-temporal classifier}
Our proposed L-TAE temporal encoder is meant to be learned alongside a spatial encoding module and a decoder module in an end-to-end fashion \eqref{eq:end2end}.
The spatial encoder $S$ maps a sequence of raw inputs $X^{(t)}$ to a sequence of learned features $e^{(t)}$, computed independently at each position of the sequence.
The decoder $D$ maps the output $o$ of the L-TAE to a target vector $y$, such as class logits in the case of a classification task.
\begin{equation}
  \left[X^{(t)}\right]_{t=1}^{T} 
  \xmapsto{\quad S \quad }
  \left[e^{(t)}\right]_{t=1}^{T}
  \xmapsto{\;\; \LTAE \;\;}
  \;\;o \;\;
  \xmapsto{\quad D \quad }
  y~.
  \label{eq:end2end}
\end{equation}
%

%% file: experiments.tex
\subsection{Dataset}
We evaluate our proposed method with the public dataset \emph{Sentinel2-Agri} \cite{garnot2019satellite}, comprised of $191\,703$ sequences of $24$ superspectral images of agricultural parcels from January to October. The acquisitions have a spatial resolution of $10$m per pixel and $10$ spectral bands. Each parcel is annotated within a $20$ class nomenclature of agricultural crops.
\subsection{Metric and Protocol}
We use two classification metrics to assess the performance of predictions: the Overall Accuracy (OA) and the mean Intersection-over-Union (mIoU). The former accounts for the precision of the prediction regardless of the class distribution, while the latter computes the IoU for each class and averages the results over the class set. Given that the dataset is unbalanced ($4$ classes represent $90\%$ of the samples) the mIoU gives a more faithful assessment of the performance. 

We propose two evaluation protocols to assess the efficiency of our proposed light-weight temporal attention encoder:
\begin{itemize}
    \item[$\bullet$] We assess the performance of our method and several state-of-the-art parcel classification algorithms on the dataset Sentinel2-Agri.
    In order to perform a fair comparison, we chose configurations corresponding to around $150$k parameters for all methods. We report the results in \tabref{tab:perf_classif} alongside the theoretical number of floating point operations (in FLOPs) required for the sequence embedding modules to process a single sequence at inference time.
    \item[$\bullet$]
    We complement this first experiment by comparing the performance of  different configurations of sequence embedding algorithms, and plot the performance with respect to the number of parameters. 
    In order to remove the effects of the different spatial encoders, we use the same spatial encoder (a pixel set encoder \cite{garnot2019satellite}) in all experiments. We only adapt the last linear layer of the spatial encoder to produce embeddings of the desired dimensions.
\end{itemize}
\subsection{Evaluated Methods}
We evaluate the performance of recent algorithms operating on satellite image time series in order to assess the relative improvement offered by our proposed method.
\begin{itemize}
    \item\textbf{PSE+TAE} 
    The approach proposed by Garnot \etal \cite{garnot2019satellite}. They use a Pixel-Set Encoder (PSE) module to encode each image independently, and process the resulting sequence of embeddings with a TAE module. The decoder $D$ is a 2-layer MLP.
    \item\textbf{PSE+L-TAE} Our proposed method. We keep the same architecture as the PSE+TAE, and replace the TAE by our L-TAE network.
    \item\textbf{CNN+GRU} 
    A similar approach \cite{garnot2019time} to {PSE+TAE}, with a CNN instead of the PSE and a Gated Recurrent Unit \cite{chung2014GRU} instead of the TAE.
    \item\textbf{CNN+TempCNN} 
    Another variation of this architecture, with a two-dimensional CNN to encode the images and a one-dimensional CNN processing the temporal dimension independently.
    This architecture is based on the work of Pelletier \etal \cite{pelletier2019temporal}.
    \item\textbf{Transformer}
    Ru{\ss}wurm \etal were the first to introduce self-attention methods to the classification of remote sensing images. In their work\cite{russwurm2019self}, the statistics of images is simply averaged over the parcels' pixels, while the resulting sequence is processed by a Transformer network \cite{vaswani2017attention}. The output sequence of embeddings is max-pooled along the temporal dimension to produce a single embedding for the input sequence.
    \item\textbf{ConvLSTM} 
    Ru{\ss}wurm \etal \cite{russwurm2018convolutional} combine the embedding of the spatial and temporal dimensions by using a ConvLSTM network \cite{xingjian2015convolutional}. This work has been adapted to process parcels instead of pixels \cite{garnot2019satellite}.
    \item\textbf{Random Forest}
    We use the temporal concatenation scheme of Bailly \etal to train a random forest of $100$ trees using the parcel-wise mean and standard deviation of the spectral bands. 
\end{itemize}
\subsection{Analysis}
In \tabref{tab:perf_classif}, we report the performances of competing methods (taken from \cite{garnot2019satellite}) and the L-TAE architecture, all obtained with a $5$-fold cross-validation scheme.
Our proposed L-TAE architecture outperforms other methods on this dataset both in overall accuracy and mIoU. 
While the OA is essentially unchanged compared to the TAE, the increase of $0.8$ mIoU points is noteworthy since our model is not only simpler but also less computationally demanding by almost an order of magnitude.

We would like to emphasize that FLOP counts do not necessarily reflect the computational speed  of the model in practice. In our non-distributed implementation, the total inference times are dominated by loading times and the spatial embedding module. However, this metric serves to illustrate the simplicity and efficicency of our network.
%
%
\begin{table}[h!]
\caption{Performance of our proposed models and competing approaches parameterized to all have $150$k parameters approximately. MFLOPs is the number of floating points operations (in $10^6$FLOPs) \emph{in the temporal feature extraction module} and for one sequence. This only applies to networks which have a clearly separated temporal module.}
\begin{center}
\begin{tabular}{lcccccc}
 &\phantom{abc}& OA   &\phantom{abc}&mIoU & \phantom{abc} &MFLOPs   \\\toprule
PSE+L-TAE (ours) && \textbf{94.3}  \footnotesize{\rpm 0.2}  & &\textbf{51.7} \footnotesize{\rpm 0.4} && \bf 0.18\\
PSE+TAE \cite{garnot2019satellite}                   & &94.2 \footnotesize{\rpm 0.1}  &&  50.9  \footnotesize{\rpm 0.8}    && 1.7 \\ 
CNN+GRU  \cite{garnot2019time}                  && 93.8 \footnotesize{\rpm 0.3}    &&   48.1  \footnotesize{\rpm 0.6} && 3.6   \\
CNN+TempCNN \cite{pelletier2019temporal}        && 93.3\footnotesize{\rpm 0.2}    &&   47.5\footnotesize{\rpm 1.0} && 0.81\\ 
Transformer \cite{russwurm2019self}                    &&  92.2 \footnotesize{\rpm 0.3 }  &&  42.8  \footnotesize{\rpm  1.1} && 1.1\\
ConvLSTM  \cite{russwurm2018convolutional}     & & 92.5\footnotesize{\rpm 0.5}    &&   42.1\footnotesize{\rpm 1.2} && - \\
Random Forest \cite{bailly2018crop}        && 91.6\footnotesize{\rpm 1.7}   && 32.5  \footnotesize{\rpm 1.4} && - \\\bottomrule
\end{tabular}
\end{center}
\vspace{-.39cm}
\label{tab:perf_classif}
\centering
\end{table}

\begin{figure}[]
\includegraphics[width=\linewidth]{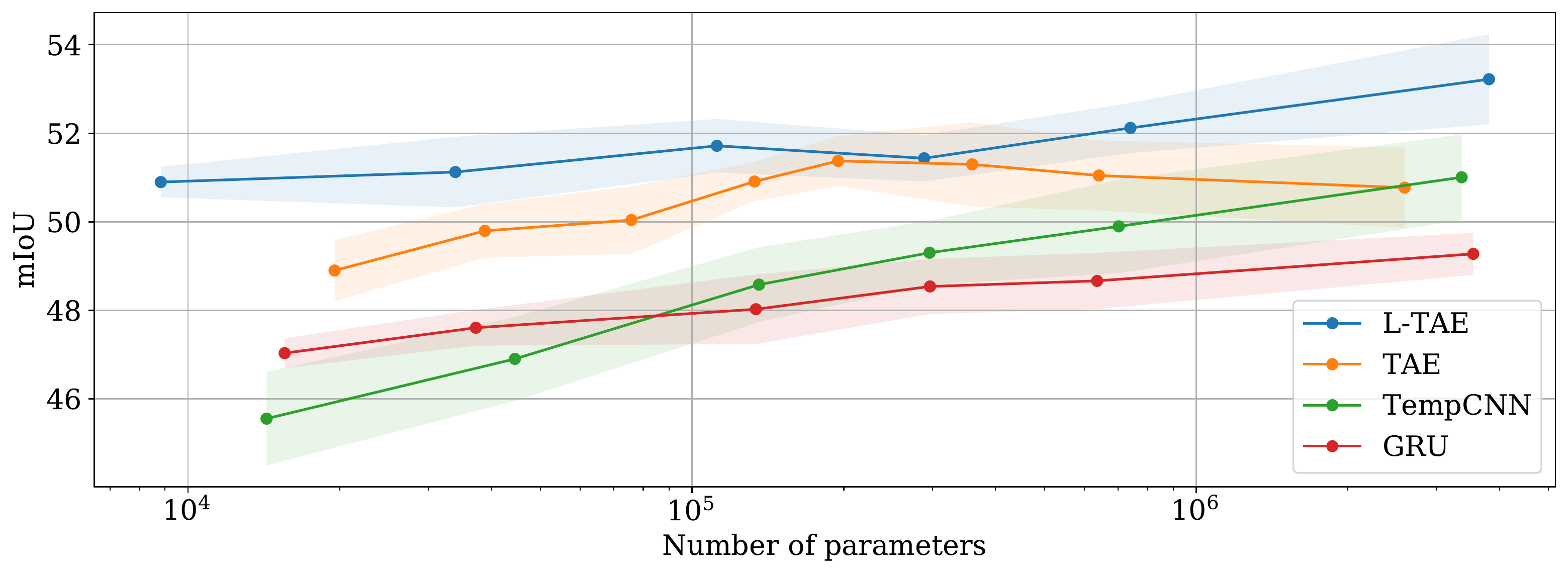}
\caption{Performance (in mIoU) of different approaches plotted with respect to the number of parameters in the sequence embedding module. The number of parameters is given on a logarithmic scale. The shaded areas depict the observed standard deviation of mIoU across the five cross-validation folds. The L-TAE outperforms other models across all model sizes, and the smallest 9k-parameter L-TAE instance yields better mIoU than the 100k-parameter TAE model.  
}
\label{fig:perfparam}
\end{figure}

Furthermore, our network maintains a high precision even with a drastic decrease in the parameter count, as illustrated in \figref{fig:perfparam}.
We evaluate the four best performing sequence embedding modules (L-TAE, TAE, GRU, TempCNN) in the previous experiment with different configurations, ranging from $9k$ to $3M$ parameters. These algorithms all operate with the same decoder and spatial module: a PSE and decoder layer totaling $31$k parameters.
The smallest L-TAE configuration, with only $9k$ parameters, achieves a better mIoU score than a TAE with almost $110k$ parameters, a TempCNN with over $700k$ parameters, and a GRU with $3M$ parameters. See \tabref{tab:all_config} in the Appendix for the detailed configurations  corresponding to each points.

\begin{figure}
    \centering
    \includegraphics[width=\linewidth]{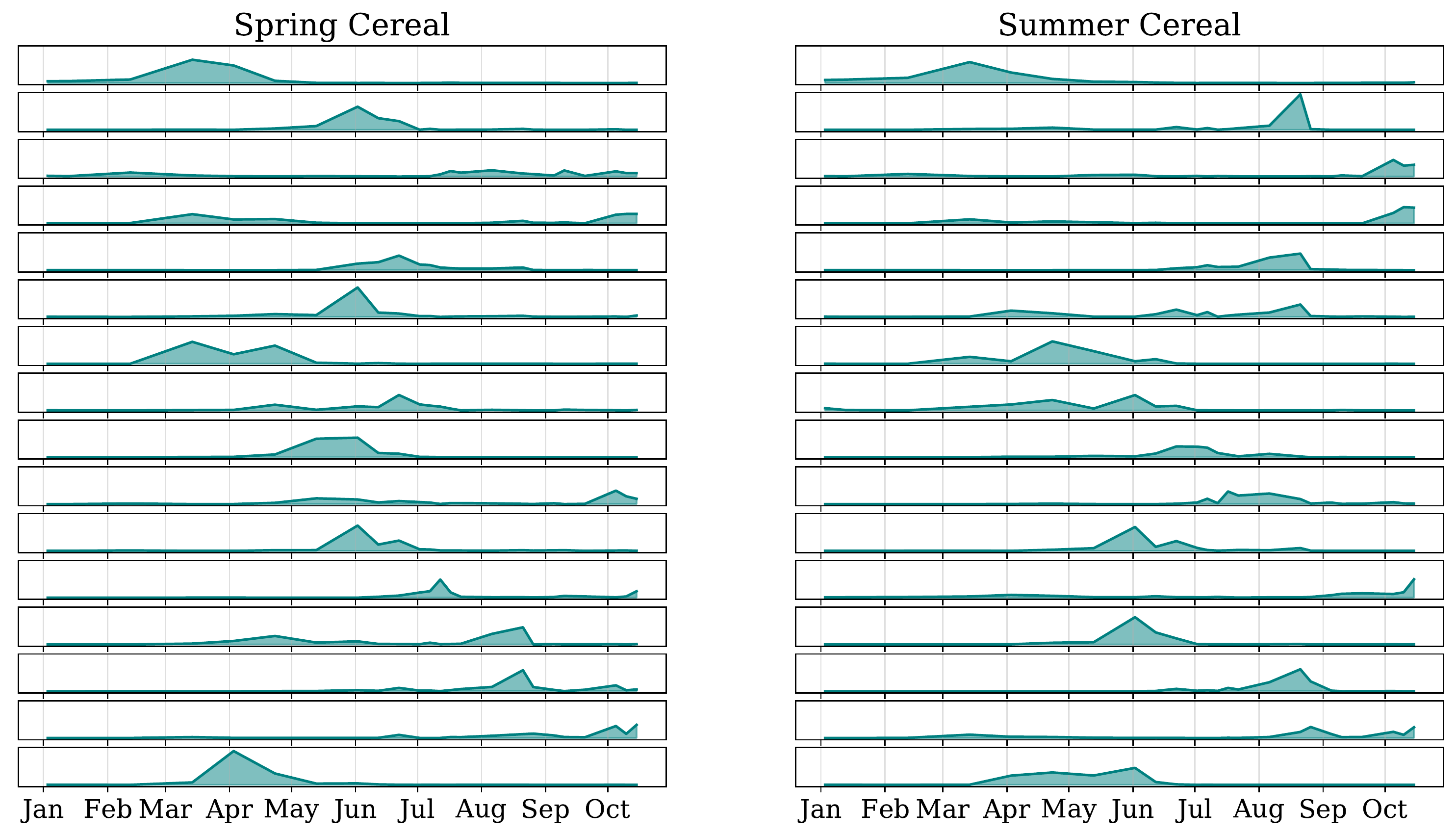}
    \caption{Average attention masks of the L-TAE for parcels of classes Spring Cereal (left) and Summer Cereal (right), for a model with $16$ heads (from top to bottom). The masks illustrate how each head focuses on short temporal intervals which depend on crop type. 
    }
    \label{fig:masks}
\end{figure}

In \figref{fig:masks}, we represent the average attention masks of a $16$-head L-TAE for two different classes. We observe that the masks of the different heads focus on narrow and distinct time-extents, \ie display a high degree of specialization. We also note that the masks are adaptive to the parcels crop types. This suggests that  the attention heads are able to cater the learned features to the plant types considered. We argue that our channel grouping strategy, in which each head processes distinct time-stamped features, allows for this specialization and leads to an efficient use of the trainable parameters.

\subsection{Ablation Study and Robustness Assessment}
 In \tabref{tab:ablations}, we report the performance of our proposed L-TAE architecture with different configurations of the following hyper-parameters: number of heads $H$, dimension of keys $K$, and number of channels $E$ in the input sequence. We note that our model retains a consistent performance throughout all configurations.
%
\paragraph{Number of heads:} 
The number of heads seems to only have a limited effect on the performance. We hypothesize that while a higher number of heads $H$ is beneficial, a smaller group size $E'$ is however detrimental. 
\if 1 0
A surprising result is the good performance achieved by the two-headed version of LTAE (51.6 mIoU). 
In \figref{fig:2heads}, we represent the attention masks of this model for different crop types. We observe that the attention masks are non longer adaptive to the nature of the input sequence. 
We hypothesize that a reason that this lack of specialization only has limited effects on precision is that all parcels of the dataset Sentinel2-Agri are taken from the same region and at the same dates, and hence share a common latent meteorological context. Experiments on a  dataset with more diverse acquisitions are needed to confirm this reasoning.
\begin{figure}
    \centering
    \includegraphics[width=\linewidth]{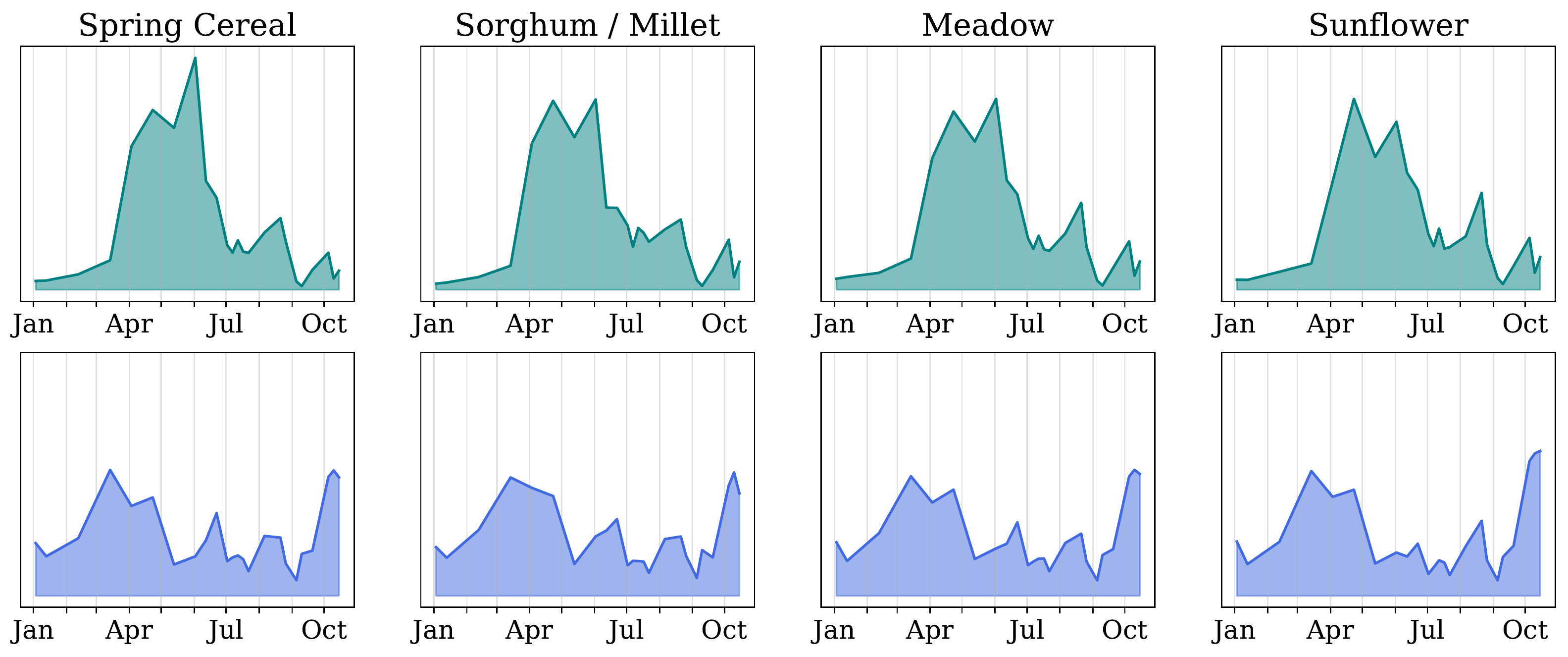}
    \caption{Average attention masks of a L-TAE with two attention heads.
    Contrarily to the 16-heads version presented in \figref{fig:masks}, the  masks are almost similar across crop types.
    }
    \label{fig:2heads}
\end{figure}
%
\fi
\paragraph{Key Dimension:} Our experiments show that smaller key dimensions than the typical values used in NLP or for the TAE ($K=32$) perform better on our problem. Even $2$-dimensional keys allow for the L-TAE to achieve performances similar to the TAE.

\paragraph{Input Dimension:} The variation in performance observed with larger input embeddings is expected: it corresponds to a richer representation. However, the returns are decreasing on the considered dataset with respect to the number of incurred parameters.

\paragraph{Query-as-Parameter}
In order to evaluate the impact of our different design choices, we train a variation of our network with the same master-query scheme than the TAE. The larger resulting linear layer increases the size of the model for a total of $170$k parameters, resulting in a mIoU of only $49.7$. This indicates that the query-as-parameter scheme is not only beneficial in terms of compactness but also performance.
%
%
\begin{table}[h]
    \caption{Impact of several hyper-parameters on the performance of our method. \underline{Underlined}, the default parameters values in this study; in \textbf{bold}, the best performance.}
    \centering
    \begin{tabular}{ccccccccccc}
    $H$ & Params.  & mIoU & \phantom{abc} & $K$ & Params.  & mIoU & \phantom{abc} &$E$ & Params.  & mIoU \\ \cmidrule{1-3} \cmidrule{5-7} \cmidrule{9-11}
     2  &  114k   &  51.6    &&      2   &     118k   &     50.7      &&  32    & 46k     &     49.6     \\
     4  &      118k   &  51.0  & & 4   &   127k     &     51.3      &&    64   &   59k   &      49.6     \\
      8&    127k  &    51.2  && \underline{8}  &   143k   &     \bf51.7     &&   128    &  65k  &      51.1     \\
    \underline{16} &    143k &   \bf51.7    && 16  &    176k    &    50.8       &&  \underline{256}     &   143k   &      \bf51.7     \\

      32 &   176k  &   51.2   &&       32  &     242k   &     51.2      &&   512    &   254k  &      51.4     \\\cmidrule{1-3} \cmidrule{5-7} \cmidrule{9-11}
    \end{tabular}
    \label{tab:ablations}
\end{table}
\subsection{Computational Complexity}
In \tabref{tab:complexity}, we report the asymptotic complexity of different sequence embedding algorithms. For the L-TAE, the channel grouping strategy removes the influence of $H$ in the computation of keys and outputs compared to a TAE or a Transformer. The complexity of the L-TAE is also lower than the GRU's as $M$, the size of the hidden state, is typically larger than $K$ ($130$ vs $8$ in the experiments presented in \tabref{tab:perf_classif}). 
\begin{table}[]
  \caption{Asymptotic complexity of different temporal extraction modules for the computation of keys, attention masks, and output vectors. For the GRU, the complexity of the memory update is given in the Keys and Mask columns. $X$ is the size of the output vector. $M$ is the size of the hidden state of the GRU.}
    \label{tab:complexity}
    \centering
\begin{tabular}{c@{\quad}c@{\quad}c@{\quad}c}
    Method & Keys &  Mask & Output \\\toprule
    L-TAE & $O(TEK)$ & $O(HTK)$&  $O(EX)$ \\
    TAE & $O(HTEK)$  & $O(HTK)$ &  $O(HEX)$ \\
    Transf. & $O(HTEK)$ & $O(HT^2K)$ & $O(HEX)$ \\
    GRU &  \multicolumn{2}{c}{$O\left(MT(E+M)\right)$} & $O(MX)$  \\\bottomrule
\end{tabular}
\end{table}

%% file: appendix.tex
In \tabref{tab:details1}, we give the exact configurations used to obtain the values in \figref{fig:perfparam}.
\begin{table}[]
\caption{Configurations of the L-TAE, TAE, GRU, and TempCNN instances used to obtain \figref{fig:perfparam}.}
\label{tab:all_config}
\label{tab:details1}
\centering
\begin{tabular}{lcccccccl}
Parameters &\phantom{ab}& E    &\phantom{ab}& H &\phantom{ab} & K &\phantom{ab}& \multicolumn{1}{c}{MLP}   \\ \toprule
\textbf{L-TAE} & & & & \\\hline
9 k        && 128  && 8  && 8 && 128                        \\
34 k        && 128  && 16 && 8 && 128 -  128                 \\
112 k       && 256  && 16 && 8 && 256 -  128                 \\
288 k       && 512  && 32 && 8 && 512 -  128                 \\
740 k       && 1024 && 32 && 8 && 1024 -  256 -  128         \\
3840 k    && 2048 && 64 && 8 && 2048 -  1024 -  256 -  128 \\\hline
\textbf{TAE} &\\\hline
19 k       && 64  && 2 && 8  && 128 -  128              \\
39 k       && 64  && 4 && 8  && 256 -  128              \\
76 k      && 128  && 4 && 8  && 512 -  128              \\
195 k      && 256  && 4 && 8  && 1024 -  128             \\
360 k      && 256  && 4 && 8  && 1024 -  256 -  128      \\
641 k      && 256  && 8 && 8  && 2048 -  256 -  128      \\
2592 k   && 1024 && 8 && 16 && 8192 -  256 -  128      
\\\bottomrule
\end{tabular}
\\~\\~\\
\centering
\begin{tabular}{lccclclcl}
Parameters &\phantom{ab}& Hidden Size & \phantom{abcdef} & Parameters & \phantom{ab}&Kernels &\phantom{ab}& FC \\\cmidrule{1-3} \cmidrule{5-9} 
15k   && 32   && 14k   && 16 -  16 -  16    && 16 -  16   \\
37k   && 64   && 45k   && 32 -  32 -  32    && 32 -  32   \\
134k  && 156  && 136k  && 64 -  64          && 64         \\
296k  && 256  && 296k  && 128 -  128        && 64         \\
636k  && 400  && 702k  && 128 -  128 -  128 && 180        \\
3545k && 1024 && 3362k && 64 -  128 -  256  && 512 -  128 \\ \cmidrule{1-3} \cmidrule{5-9} 
\multicolumn{3}{c}{\textbf{GRU}} & & \multicolumn{5}{c}{\textbf{TempCNN}}
\end{tabular}
\end{table}